\documentclass{article}

\usepackage{pythontex}
\usepackage{amsmath}

    \usepackage[final]{neurips_2023}

\usepackage[utf8]{inputenc} 
\usepackage[T1]{fontenc}    
\usepackage{hyperref}       
\usepackage{url}            
\usepackage{booktabs}       
\usepackage{amsfonts}       
\usepackage{nicefrac}       
\usepackage{microtype}      
\usepackage{xcolor}         
\usepackage{enumitem}
\usepackage{tikz}
\usepackage{pgfplots}
\usepgfplotslibrary{statistics} 
\usepgfplotslibrary{fillbetween} 
\pgfplotsset{compat=newest} 
\usetikzlibrary {arrows.meta}
\usepackage{tikzscale}
\usepackage{enumitem}

\newcommand{\Figref}[1]{Fig.~\ref{#1}}
\newcommand{\Tabref}[1]{Tab.~\ref{#1}}
\newcommand{\Secref}[1]{Sec.~\ref{#1}}

\newcommand{\hide}[1]{}

\setcitestyle{square}

\title{Scaling TabPFN: Sketching and Feature Selection for Tabular Prior-Data Fitted Networks}

\author{Benjamin Feuer, Chinmay Hegde, Niv Cohen \\
\vspace{3pt}
New York University \\
\vspace{3pt}
\{bf996, chinmay.h, niv.cohen\}@nyu.edu}

\begin{document}

\maketitle

\begin{abstract}
Tabular classification has traditionally relied on supervised algorithms, which estimate the parameters of a prediction model using its training data. Recently,  Prior-Data Fitted Networks (PFNs) such as TabPFN have successfully learned to classify tabular data \textit{in-context}: the model parameters are designed to classify new samples based on labelled training samples given \textit{after} the model training. While such models show great promise, their applicability to real-world data remains limited due to the computational scale needed.
Here we study the following question: given a pre-trained PFN for tabular data, what is the best way to summarize the labelled training samples before feeding them to the model? We conduct an initial investigation of sketching and feature-selection methods for TabPFN, and note certain key differences between it and conventionally fitted tabular models. 
\end{abstract}

\section{Introduction} 
\vspace{-2mm}

Classification of tabular data is a basic machine learning task of vital importance, and has inspired a large set of algorithmic approaches. These approaches, ranging from classical algorithms such as gradient-boosted trees~\citep{prokhorenkova_catboost_2018, xgboost} to recent attempts with deep learning~\citep{somepalli2021saint} have all been focused on choosing proper model parameters (and hyper-parameters) given an explicitly defined model hypothesis class. 

In a recent parallel line of work, \emph{Prior-Data Fitted Networks} such as TabPFN have successfully been demonstrated to classify tabular data based on a training set given as the model \textit{input}~\citep{hollmann_tabpfn_2023}. Rather than using training data to fit the model parameters, TabPFN gets at inference time a set that contains \textit{both} labeled and unlabeled samples. It then predicts the `missing' labels directly based on the labeled input samples, rather than exclusively relying on trained model parameters. In this sense, the working of TabPFN resembles the phenomenon of \emph{in-context learning} that is exhibited by large language models such as GPT, which emerges during pretraining~\citep{brown_language_2020, li2023emergent}. In TabPFN, the pretraining stage can be viewed as inducing suitable statistical priors which are then used to fill in the missing labels.


In large language models, the choice (and length) of the context crucially affects prediction performance~\citep{xie_explanation_2022}. Optimizing the performance of large language models often requires long (and fairly complicated) prompting strategies. Many recent studies have examined how prompting in-context learning algorithms allows the model to retrieve memorized information, perform simple mathematical operations, or even execute high-level reasoning \citep{liu_goat_2023, meng2022locating, nanda_progress_2023}. Context-optimization strategies have been  less frequently studied for tabular models such as TabPFN. They are, however, crucial, as memory constraints prevent TabPFN from using more than a few thousand samples as context.

When using Prior-Data Fitted Networks for tabular data, the ``prompt'' is composed from the sample values and their labels from the training set. The model is then expected to give predictions on a set of unlabelled samples, also given as part of the prompt. This setting is simpler than the case for language models: all the labelled parts of the prompt are weighed equally, and the labeled (training) samples can be treated as an unordered set, rather than parts of an ordered string with internal syntax.

Focusing on in-context learning for tabular data has additional advantages: (i) While the evaluation of language models is a hard problem (covering many tasks and complicated metrics) the evaluation of tabular data classifiers is straightforward. (ii) The prior knowledge implicitly contained in the pretrained model is driven in the tabular case by synthetic data, unlike the case with large language models which use large web-scraped text corpora. (iii) Improving the results of TabPFN further advances the use of deep models for tabular data and may be of practical importance. 

Here, we focus on understanding one part of the context optimization problem for tabular classification: summarizing a large training dataset, $X_\text{labelled}$, into a more compact set, $X_\text{compact}$, such that most of the information useful to the model is contained in $X_\text{compact}$. The relative simplicity of tabular in-context learning allows us to formalize a concrete question:
Given a pretrained model, which operations should be used to summarize the input target data before feeding it to TabPFN? 

In this short paper, we empirically study basic properties about how one should (or should not) summarize a target tabular dataset when used as context for in-context learning.




\section{Sketching and feature selection for tabular in-context learning}
\vspace{-2mm}

Feature selection and sketching methods have been extensively explored in prior literature on tabular classification~\citep{munteanu_coresets-methods_2018, chandrashekar_survey_2014}. In the following section, we report the results of applying a representative selection of these methods.

\subsection{Experimental setting}
\label{sec:setting}
\vspace{-2mm}

To systematically evaluate TabPFN with the different context summarization we select a subset of nineteen datasets from \citep{mcelfresh_when_2023} which exceed either the feature or sample limitation recommended by the authors of TabPFN, which are 100 features and 1000 samples, respectively~\citep{hollmann_tabpfn_2023}. We limit our algorithmic comparison to TabPFN and CatBoost, which is the best-performing overall model in \citep{mcelfresh_when_2023}. The complete list of datasets can be found in \Secref{app:dataset-list}.
For ease of comparison with existing meta-analyses, where possible we replicate the method of \citep{mcelfresh_when_2023}. We compare CatBoost with 30 hyperparameter settings (one default set and 29 random sets, using Optuna) to TabPFN, averaging over 10 train/validation folds for each dataset.

Our main results can be found in \Tabref{tab:main-results}. We conduct our primary investigation into sketching methods at $d_\text{max} = 100$ features and $n_\text{max} = 3000$ samples from each dataset. Dataset names are drawn from OpenML with abbreviations as follows: gddc refers to gas-drift-different-concentrations, fm to Fashion-MNIST, ss to skin-segmentation. CB stands for CatBoost, TP for TabPFN, f for full dataset, b for best result using any combination of algorithms, and r for the best result using random feature and random sample selection. SKT is an abbreviation for sketching method, FTS for feature selection, SMP for sampling strategy. We report the most successful combination for each model, on each dataset, with respect to average accuracy over ten folds.

We determine statistically significant ($p < 0.05$) performance differences between algorithms by use of a Wilcoxon signed-rank test between CatBoost and TabPFN. A Holm-Bonferroni correction is used to account for multiple comparisons.

\begin{table}[]
\resizebox{\columnwidth}{!}{%
\begin{tabular}{@{}lrrrrrll@{}}
\toprule
\textbf{Dataset} & \multicolumn{1}{l}{\textbf{Acc (CB, f)}} & \multicolumn{1}{l}{\textbf{Acc (CB, b)}} & \multicolumn{1}{l}{\textbf{Acc (CB, r)}} & \multicolumn{1}{l}{\textbf{Acc (TP, b)}} & \multicolumn{1}{l}{\textbf{Acc (TP, r)}} & \textbf{SKT / FTS / SMP (CB)} & \textbf{SKT / FTS / SMP (TP)} \\ \midrule
airlines\_189354 & \textbf{0.653} & 0.637 & 0.637 & 0.594 & 0.589 & RND / RND / PR & RND / RND / PR \\
albert\_189356 & \textbf{0.698} & 0.657 & 0.657 & 0.64 & 0.64 & RND / RND / PR & RND / RND / PR \\
CIFAR\_10\_167124 & \textbf{0.434} & 0.37 & 0.342 & 0.373 & 0.372 & RND / PCA / PR & RND / RND / PR \\
connect-4\_146195 & \textbf{0.749} & 0.716 & 0.716 & 0.66 & 0.659 & RND / RND / PR & RND / RND / PR \\
eeg-eye-state\_14951 & 0.832 & 0.808 & 0.806 & \textbf{0.932} & \textbf{0.932} & RND / RND / PR & RND / RND / EQ \\
elevators\_3711 & 0.855 & 0.838 & 0.838 & \textbf{0.9} & \textbf{0.899} & RND / MUT / PR & RND / RND / PR \\
FM\_146825 & \textbf{0.843} & 0.787 & 0.787 & \textbf{0.835} & 0.812 & RND / RND / PR & RND / PCA / PR \\
gddc\_9987 & 0.97 & 0.976 & 0.955 & \textbf{0.994} & \textbf{0.993} & RND / PCA / EQ & RND / RND / PR \\
higgs\_146606 & \textbf{0.71} & 0.684 & 0.684 & 0.665 & 0.661 & RND / RND / PR & RND / RND / PR \\
hill-valley\_145847 & 0.514 & 0.514 & 0.514 & \textbf{0.56} & \textbf{0.56} & RND / RND / PR & RND / RND / PR \\
mfeat-factors\_12 & 0.954 & 0.95 & 0.943 & \textbf{0.973} & \textbf{0.973} & KMN / RND / EQ & RND / RND / PR \\
mfeat-pixel\_146824 & 0.955 & 0.951 & 0.951 & \textbf{0.971} & \textbf{0.97} & RND / RND / PR & RND / RND / PR \\
pendigits\_32 & 0.972 & 0.966 & 0.964 & \textbf{0.995} & \textbf{0.993} & RND / RND / PR & RND / RND / PR \\
poker-hand\_9890 & \textbf{0.664} & 0.572 & 0.561 & 0.519 & 0.515 & RND / RND / PR & RND / RND / PR \\
riccardo\_168338 & 0.951 & 0.956 & 0.93 & \textbf{0.991} & 0.982 & RND / PCA / EQ & RND / MUT / EQ \\
robert\_168332 & \textbf{0.446} & 0.367 & 0.367 & 0.384 & 0.359 & RND / RND / PR & RND / PCA / EQ \\
semeion\_9964 & 0.887 & 0.869 & 0.863 & \textbf{0.915} & \textbf{0.915} & RND / MUT / EQ & RND / RND / PR \\
ss\_9965 & \textbf{0.994} & 0.989 & 0.987 & \textbf{0.999} & \textbf{0.999} & RND / RND / PR & RND / RND / PR \\
volkert\_168331 & \textbf{0.608} & 0.56 & 0.56 & 0.557 & 0.555 & RND / RND / PR & RND / RND / PR \\ \bottomrule
\end{tabular}%
}
\caption{\sl\textbf{Comparative performance of TabPFN (TP) and CatBoost (CB) with sketching methods. } \sl  We compare at a fixed feature size of 100 and a fixed sample size of 3000. When both models are limited to 3000 samples, TabPFN performs better on 12 of 17 datasets where significant differences exist. When Catboost is allowed access to the entire training data, the win rate is identical. In most cases, random sample selection is sufficient for optimal performance. Both models benefit from PCA and mutual information dimensionality reduction when the feature space is large.
 \textbf{Bold} indicates the best-performing model(s).}
\label{tab:main-results}
\vspace{-4mm}
\end{table}

\subsection{The effect of scale for tabular data classification}
\label{sec:exp_motivation}
\vspace{-2mm}

We begin with a short empirical study of the effect of the number of supplied labelled samples on the accuracy of our compared algorithms (TabPFN and CatBoost), noting the following interesting facts: 

(i) Although the authors of TabPFN do not recommend using the model beyond its 1000-sample context length limit \citep{hollmann_tabpfn_2023} without retraining, using larger context lengths can lead to significant gains. In \Figref{fig:feature_sample_scaling}, we ablate the effect of using different quantities of samples. We consider 100, 500, and 3000 samples for both CatBoost and TabPFN. Whiskers represent one standard deviation from the mean. Feature subsets are taken using mutual information, sampling is random, and class weighting is proportional.

(ii) Performance improves on most datasets as context length increases from 100 to 3000 samples, but the gains are far more pronounced in CatBoost than in TabPFN. 

(iii) TabPFN often outperforms CatBoost at sample sizes up to 1000, and remains highly competitive above that threshold. That said, we acknowledge that with a more rigorous parameter search, either model may be capable of exceeding our reported metrics (achieved using random hyperparameter sweeps).

\subsection{Sketching for tabular in-context learning}
\vspace{-2mm}

\label{sec:sketching}
Context length often limits the usefulness of TabPFN (Fig.\ref{fig:feature_sample_scaling}). Therefore, when reducing the given number of samples to a smaller subset, we wish the subset to preserve as much useful context from the original dataset. Since the utilization of context by the model $T$ is not explicitly known, we turn to empirical investigation. The summarization of context samples may depend on the samples themselves $X_\text{labelled}$, their labels $y$, or both. We study these factors independently: we examine summarization methods for $X_\text{labelled}$, and apply them according to the labels $y$ in one of two manners: \textit{equal}: having a similar amount of samples from each $y$ label; and \textit{proportion}:  keeping the number of samples from each label proportional to their abundance in the original data. 


In terms of the samples summarization method, we investigate a few options: \textit{random:} picking a random subset of samples. \textit{K-means:} Choosing the samples as the K-means cluster centers of the original data, where $K$ is determined according to $n_\text{max}$. \textit{CoreSet \cite{agarwal2005geometric}:} Choosing an $n_\text{max}$ sized CoreSet. We implement our methods using the \texttt{faiss} library.~\citep{johnson2019billion}

\textbf{Results.}
We summarize the results in \Tabref{tab:main-results}. 

A key takeaway is that in all our experiments except one, we find that random sub-selection of samples works as well as any other method, making it a strong baseline for future experiments. 

CatBoost does show a statistically significant benefit from KMeans sketching on the \textit{mfeat-factors} dataset. Intriguingly, \textit{mfeat-factors} only has 2000 samples in the dataset; it is probable that sampling with replacement using the equal strategy leads to a better sample emphasis with $K$-means, compared to random sampling (by having a less biased model).

Sampling $y$ values in an equal (vs. proportional) manner is beneficial on 21\% of datasets when using CatBoost, and on 16\% of datasets when using TabPFN. In some cases, the difference is quite large; the best context-summarization for TabPFN / \textit{riccardo} using equal sampling attained an accuracy of $99$\%, compared to just $80$\% when using proportional sampling.

\subsection{Feature dimensionality reduction for tabular in-context learning}
\label{sec:dim_reduction}
\vspace{-2mm}

We investigate three options for feature dimensionality reduction: \textit{random reduction:} picking a random subset of features. \textit{mutual information: } selects features with high mutual information to the target dataset \citep{vergara2014review}. \textit{PCA:} taking the $d_{\max}$ first principal components. For the latter we use the scikit-learn implementations~\citep{scikit-learn}.


We find that feature subsampling often has a significant effect on classifier performance, and that the in-context classification method of TabPFN is more dependent on feature selection than that of CatBoost.
4 of our datasets have more than 256 features; (\textit{riccardo}, \textit{Robert}, \textit{Fashion-MNIST}, \textit{CIFAR-10}). 
For \textit{Robert}, \textit{FashionMNIST}, and \textit{riccardo}, the best TabPFN setting with mutual information or PCA feature selection outperforms random reduction. See \Tabref{tab:main-results}.

\begin{figure*}[h]
    \begin{tabular}{c c}
    \resizebox{0.47\textwidth}{!}{
    \includegraphics{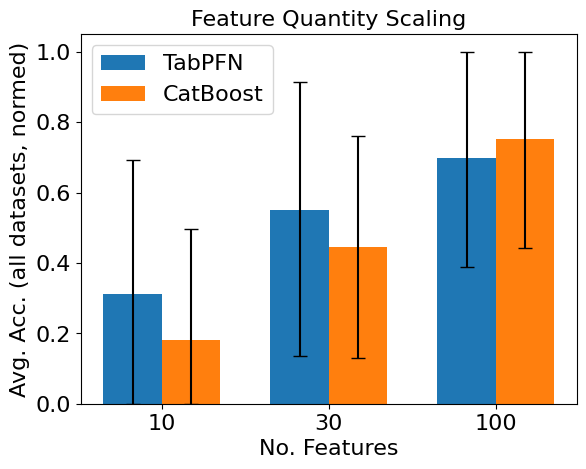}} & 
    \resizebox{0.47\textwidth}{!}{
    \includegraphics{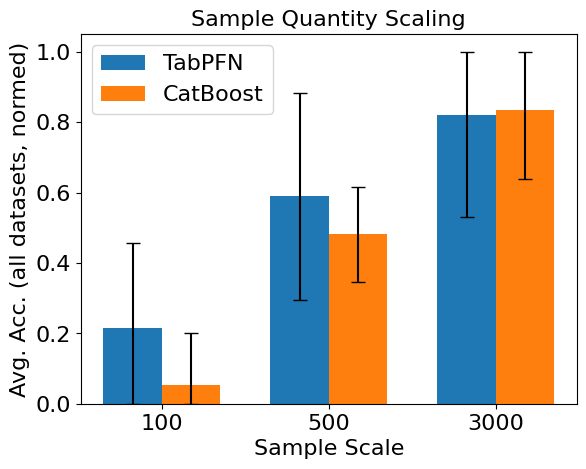}}
    \end{tabular}
    
    \caption{\sl\textbf{(L) Effects of feature quantity scaling. (R) Effects of sample quantity scaling.} (L) While both models benefit from feature quantity scaling from 10 to 100 features seen, the gains are more pronounced and consistent in CatBoost. (R) Performance gains in TabPFN are more gradual than in CatBoost as sample quantity increases from 100 to 3000 training samples seen, and less consistent. (Both) The y-axis represents the average normalized accuracy across all datasets. In the case that a given dataset has fewer samples or features than our quantity scaling figure, we take all samples or features from that dataset. The error bars represent one standard deviation from the mean. Image best viewed in color.}
    \label{fig:feature_sample_scaling}
\vspace{-4mm}
\end{figure*}

On a related note, we report in \Figref{fig:feature_sample_scaling} the mean normalized performance for both TabPFN and CatBoost with 10, 30, and 100 features. Whiskers represent one standard deviation from the mean. Feature subsets are taken using mutual information, sampling is random, and class weighting is proportional. While both models improve with more samples, the effect is more pronounced and consistent in CatBoost; as feature quantity scales up, performance is the same or better on all datasets. When we reduce the feature space in TabPFN from 100 to 10 features, performance \textit{improves} on two datasets (\textit{higgs} and \textit{connect-4}). This indicates that in-context tabular classification may be more sensitive to the presence of spurious features than supervised methods.


\section{Related works}
\label{sec:related_works}

\textbf{Evaluation of deep learning for tabular data.}\citep{deepnns} introduced into the literature TabSurvey, an exhaustive comparison between existing deep learning approaches for heterogeneous tabular data. Their work was extended recently by \citep{mcelfresh_when_2023}. Like our work, the latter directly compares CatBoost and TabPFN with sample and feature sketching. We extend this considerably, demonstrating the effects of data scaling and the comparative importance of feature and sample selection methods.

\textbf{Active learning.} Another relevant line of works considered selecting an optimal subset of a dataset to be labeled \cite{hacohen2022active,yehuda2022active,sener2017active}. While this field mostly looks to ease the labeling burden, rather than to save computational resources, it also looks to effectively reduce the number of samples included in the training set.

\textbf{Efficient attention mechanisms for transformers.} The runtime and memory usage of transformer architectures such as TabPFN grows quadratically with the number of training samples, placing severe computational bounds on their ability to scale. A wide range of efficient attention mechanisms have been proposed as a workaround for this limitation.~\cite{press2021train,chen2023extending,sun2022length, xiao_efficient_2023, han_hyperattention_2023} We consider such mechanisms an important direction for future research into scaling TabPFN.  

\hide{
\textcolor{green}{\textbf{Context optimization in large language models.} The computational scale required by large transformer models has been a significant bottleneck in natural language processing, inspiring many works on this topic \cite{}.  In large language models, the main challenge is scaling the attention mechanism during training, driving methods that train with a limited context and extend the length of used context during inference \cite{press2021train,chen2023extending,sun2022length}. Here, we focus on allowing a larger used context during inference, regardless of training complexity. Yet, carrying ideas from large language models context optimization may prove useful, including optimizing the training procedure of TabPFN for longer contexts.}

 ALiBi (Press et al., 2022) and LeX (Sun et al., 2022)
}

\textbf{Feature selection.} Other related fields study a similar problem to ours for other reasons. Feature selection and feature extraction aim to select a subset of features (or combinations of them) to achieve various goals. Such goals include building simpler, faster, and more comprehensible models \citep{li2017feature,kumar2014feature,zebari2020comprehensive}. Our initial study on feature selection or extraction for TabPFN and the examined methods fall well within the methods in these fields of study; suggesting a new application to it. We expect the nature of this new application, to highlight new feature selection techniques, which might differ from the optimal techniques used for other goals.

\textbf{Data valuation.}
The field of data valuation aims to quantify the economic value of different parts of a dataset to fairly compensate different data contributors \cite{jia2019towards,sim2022data,Fleckenstein2023Review}. Here, our focus is on giving the optimal context for a predictor, rather than understanding the fair value of each individual part.

\section{Conclusions}

Taken together, our results show that TabPFN can be scaled to perform competitively with CatBoost. First, in terms of the number of used samples, although smart representative selection methods do not boost the results significantly, simply selecting a larger subset than the one used by the authors leads to significant gains. Second, TabPFN can benefit considerably from careful feature selection, more so than CatBoost. We also believe that finding further connections between increasing the effective context of tabular prior fitted networks and optimizing the context of language in-context-learning is an exciting future research direction.


{
\bibliographystyle{neurips_2023}
\bibliography{neurips_2023}
}

\clearpage

\section{Discussions}

\textit{Why is feature dimensionality reduction more impactful than sketching?} While a complete rigorous investigation of this result is left for future work, we do note that the sample size is typically very large (e.g., $3000$) even after sketching, while the feature dimension is often reduced from a few thousands to merely $100$ features. We hypothesis that a very large set size allows some sort of statistical convergence in the sample dimension, not occurring when sub-sampling a large feature space to $100$ features in a naive way.

\section{Complete list of datasets}
\label{app:dataset-list}

In \Tabref{tab:datasets}, we list some key features of the nineteen datasets included in our meta-analysis. Additional information on the datasets is available from the OpenML website.

\begin{table}[h]
\centering
\resizebox{0.95\linewidth}{!}
{
\begin{tabular}{@{}lrrrr@{}}
                                         & \multicolumn{1}{l}{}                    & \multicolumn{1}{l}{}                     & \multicolumn{1}{l}{}                    & \multicolumn{1}{l}{}                                  \\
                                         & \multicolumn{1}{l}{}                    & \multicolumn{1}{l}{}                     & \multicolumn{1}{l}{}                    & \multicolumn{1}{l}{}                                  \\
                                         & \multicolumn{1}{l}{}                    & \multicolumn{1}{l}{}                     & \multicolumn{1}{l}{}                    & \multicolumn{1}{l}{}                                  \\
                                         & \multicolumn{1}{l}{}                    & \multicolumn{1}{l}{}                     & \multicolumn{1}{l}{}                    & \multicolumn{1}{l}{}                                  \\
                                         & \multicolumn{1}{l}{}                    & \multicolumn{1}{l}{}                     & \multicolumn{1}{l}{}                    & \multicolumn{1}{l}{}                                  \\
                                         & \multicolumn{1}{l}{}                    & \multicolumn{1}{l}{}                     & \multicolumn{1}{l}{}                    & \multicolumn{1}{l}{}                                  \\
                                         & \multicolumn{1}{l}{}                    & \multicolumn{1}{l}{}                     & \multicolumn{1}{l}{}                    & \multicolumn{1}{l}{}                                  \\
                                         & \multicolumn{1}{l}{}                    & \multicolumn{1}{l}{}                     & \multicolumn{1}{l}{}                    & \multicolumn{1}{l}{}                                  \\
\textbf{dataset}                                & \multicolumn{1}{l}{\textbf{n. classes}} & \multicolumn{1}{l}{\textbf{n. features}} & \multicolumn{1}{l}{\textbf{n. samples}} & \multicolumn{1}{l}{\textbf{pct. samples seen @ 3000}} \\
\midrule
airlines\_189354                         & 2                                       & 7                                        & 539383                                  & 0.6                                                   \\
albert\_189356                           & 2                                       & 78                                       & 425240                                  & 0.7                                                   \\
CIFAR\_10\_167124                        & 10                                      & 3072                                     & 60000                                   & 5                                                     \\
connect-4\_146195                        & 3                                       & 42                                       & 67557                                   & 4.4                                                   \\
eeg-eye-state\_14951                     & 2                                       & 14                                       & 14980                                   & 20                                                    \\
elevators\_3711                          & 2                                       & 18                                       & 16599                                   & 18.1                                                  \\
Fashion-MNIST\_146825                    & 10                                      & 784                                      & 70000                                   & 4.3                                                   \\
gas-drift-different-concentrations\_9987 & 6                                       & 129                                      & 13910                                   & 21.6                                                  \\
higgs\_146606                            & 2                                       & 28                                       & 98050                                   & 3.1                                                   \\
hill-valley\_145847                      & 2                                       & 100                                      & 1212                                    & 100                                                   \\
mfeat-factors\_12                        & 10                                      & 216                                      & 2000                                    & 100                                                   \\
mfeat-pixel\_146824                      & 10                                      & 240                                      & 2000                                    & 100                                                   \\
pendigits\_32                            & 10                                      & 16                                       & 10992                                   & 27.3                                                  \\
poker-hand\_9890                         & 10                                      & 10                                       & 1025009                                 & 0.3                                                   \\
riccardo\_168338                         & 2                                       & 4296                                     & 20000                                   & 15                                                    \\
robert\_168332                           & 10                                      & 7200                                     & 10000                                   & 30                                                    \\
semeion\_9964                            & 10                                      & 256                                      & 1593                                    & 100                                                   \\
skin-segmentation\_9965                  & 2                                       & 3                                        & 245057                                  & 1.2                                                   \\
volkert\_168331                          & 10                                      & 180                                      & 58310                                   & 5.1                                                  
\end{tabular}}
\caption{\textbf{Complete list of datasets. }}
\label{tab:datasets}
\end{table}

\section{Formal statement of our optimization problem}
For completeness, we provide here a mathematical formulation of our optimization objective.

Given a pretrained model $T$, which function $S$ should we use to scale input target data before feeding it to the model? Specifically, for an $m$-class classification problem, we wish to reduce the size of the target labelled dataset $X \in \mathcal{R}^{n\times d}$, and labels $y \in \{0,\ldots,m\}^n$ such that the sample number would be limited by $n_\text{max}$  and the feature dimension by $d_\text{max}$.
We describe the reduction of the labelled set as a $(X_\text{compact}, y_\text{compact}) = S(X_\text{labelled},y)$. We wish to find a function $S$ such that using it the unlabelled sample $X_\text{test}$ would be correctly classified by the prompted model $T$, according to their ground truth labels $t_\text{test}$ (not given during training):
\begin{align*}
\underset{S}{\max} & \Bigl(\bigl( \text{Accuracy}(T(X_\text{compact}; y_\text{compact}; X_{\text{test}}), y_{\text{test}}\bigl) \Bigl) \\
\text{s.t.  } & X_{\text{compact}} \in \mathbb{R}^{n' \times d'}, \quad d' \leq d_\text{max}, \quad n' \leq n_\text{max}.
\end{align*}

\end{document}